\documentclass[conference]{IEEEtran}
\IEEEoverridecommandlockouts
\usepackage{cite}
\usepackage{caption}
\usepackage{amsmath,amssymb,amsfonts}

\usepackage{graphicx}
\usepackage{textcomp}
\usepackage{xcolor}
\usepackage{rotating}
\usepackage{enumitem}
\usepackage{url}
\usepackage{tabularx}
\usepackage{multicol}
\usepackage{multirow}
\usepackage{subcaption}
\usepackage{threeparttable}
\usepackage{float}
\usepackage{algorithm}
\usepackage[noend]{algpseudocode}
\usepackage{makecell}
\usepackage{adjustbox}
\usepackage{placeins}
\usepackage{stfloats} 
\usepackage{booktabs}
\usepackage{rotating}
\usepackage{graphicx}
\usepackage{amsmath}
\usepackage{amssymb}
\usepackage{booktabs}
\usepackage{graphicx}
\usepackage{tabularx}
\usepackage{algorithm}
\usepackage{tabularx}
\usepackage{multicol}
\usepackage{multirow}
\usepackage{subcaption}
\usepackage{threeparttable}
\usepackage[noend]{algpseudocode}
\def\BibTeX{{\rm B\kern-.05em{\sc i\kern-.025em b}\kern-.08em
    T\kern-.1667em\lower.7ex\hbox{E}\kern-.125emX}}
\begin{document}

\title{Leveraging ChatGPT and Other NLP Methods for Identifying Risk and Protective Behaviors in MSM: Social Media and Dating apps Text Analysis
}
\author{
    \IEEEauthorblockN{
        Mehrab Beikzadeh\IEEEauthorrefmark{1},
        Chenglin Hong \IEEEauthorrefmark{2},
        Cory J Cascalheira\IEEEauthorrefmark{3},\IEEEauthorrefmark{4},\\ 
        Callisto Boka\IEEEauthorrefmark{2},
        Majid Sarrafzadeh\IEEEauthorrefmark{1},
        Ian W Holloway\IEEEauthorrefmark{2}
    }
    \IEEEauthorblockA{\IEEEauthorrefmark{1}\textit{Department of Computer Science, Henry Samueli School of Engineering,} \\
    \textit{University of California, Los Angeles, Los Angeles, CA}
    }
    \IEEEauthorblockA{\IEEEauthorrefmark{2}\textit{Department of Social Welfare, Luskin School of Public Affairs,} \\
    \textit{University of California, Los Angeles, Los Angeles, CA}
    }
    \IEEEauthorblockA{\IEEEauthorrefmark{3}\textit{Department of Counseling \& Educational Psychology,} \\
    \textit{New Mexico State University, Las Cruces, NM}
    }
    \IEEEauthorblockA{\IEEEauthorrefmark{4}\textit{VA Puget Sound Healthcare System,} \\ 
    \textit{Seattle, WA}
    }
}

\maketitle

\begin{abstract}
\textbf{Background:} Men who have sex with men (MSM) are at elevated risk for sexually transmitted infections and harmful drinking compared to their heterosexual counterparts. Text data collected from social media and dating apps may provide new opportunities for personalized public health interventions, as engagement in risk and protective behaviors could be identified automatically.\newline
\textbf{Objective:} Our objective was to determine whether social media and dating apps text data can be used to predict risk and protective behaviors among MSM.\newline
\textbf{Methods:} We gathered textual data from social media and dating apps with users’ consent and trained machine learning models to identify various risk behaviors, such as condomless anal sex, number of sexual partners, binge drinking, and heavy drinking. Drinking outcomes were based on the Alcohol Use Disorders Identification Test (AUDIT-C). We also trained a model to determine whether an individual was using pre-exposure prophylaxis (PrEP). Features were extracted from the text using ChatGPT embeddings, BERT embeddings, LIWC analysis, and a custom dictionary-based approach for identifying risk terms.\newline
\textbf{Results:} Our model was highly predictive of monthly binge drinking and having over 5 sexual partners (F1 scores 0.78 and 0.78), but slightly less predictive of taking PrEP use and heavy drinking (F1 scores 0.64 and 0.63). Overall, ChatGPT embeddings were found to be highly informative in prediction, but combining ChatGPT embeddings with LIWC and BERT and using the most correlated features improved performance on the prediction task.\newline
\textbf{Conclusions:} Our findings demonstrate that text data has the potential to provide valuable insights into specific risk and protective behaviors. By increasing the volume of data collected, we could enhance and refine the results, particularly for behaviors that were less common in the study population. Such models hold promise in guiding personalized public health interventions for MSM.\newline
\end{abstract}

\begin{IEEEkeywords}
machine learning, HIV risk, harmful drinking, social app, dating app, Text mining, ChatGPT, eHealth, LLM.
\end{IEEEkeywords}

\section{Introduction}
Over half of new human immunodeficiency virus (HIV) infections occur among men who have sex with men, primarily due to engagement in high-risk sexual behaviors and intravenous drug use\cite{Cen,Hiv}. 
Substance use is a well-established correlate of engagement in HIV risk behavior for this population\cite{Ian}
Research indicates that these health disparities in substance use and HIV are influenced by unjust social conditions\cite{Par,Sing} and increased exposure to minority stressors\cite{Bro,Mey}. MSM are more prone to mental distress and depression\cite{Gon} potentially leading to increased substance use as a coping mechanism.\cite{Wea}

Alcohol consumption, particularly binge and heavy drinking, has been identified as a significant driver of risk sexual behaviors among MSM, leading to an increased risk of HIV transmission\cite{alcoholmsm}. Notably, studies have found a higher prevalence of alcohol-related problems among MSM than among the general population of men (GPM). \cite{Med},This has further implications as the prevalence of heavy and binge drinking among MSM exacerbates the associated health risks.


However, a global survey among substance-using MSM found that only 11\% of respondents had access to substance use treatment programs and 5\% participated in such a program \cite{Flo}. In the US, only 6.5\% of people who needed substance use treatment received it in 2020 \cite{Sub}. Majority of people who were determined to need treatment did not recognize the need themselves, but among the people who wanted treatment, the main reasons for not receiving it were affordability due to the lack of health care coverage, not finding an appropriate program, and fear of others having a negative opinion of them. Mobile- and eHealth-based interventions could improve the accessibility of interventions, as they can be accessed regardless of geographic location.
Mobile and eHealth interventions offer potential for personalization by leveraging increased access to participant data. Earlier research has demonstrated successful personalized HIV interventions for MSM and individuals using substances\cite{Sul,Dill,Ing}. However, this customization often relies on participants self-reporting their behaviors, which can become burdensome. For instance, a study asked participants to complete daily or biweekly surveys, with many finding them monotonous or overly frequent, regardless of the chosen frequency\cite{Swe}. Nevertheless, the group receiving daily surveys found them more beneficial, as they better captured the frequent behavioral changes. This suggests that continuous behavior monitoring without excessively relying on participant input could be advantageous. Automating some or all of the behavior monitoring could, therefore, alleviate the burden for participants.\par
In this research, we aim to explore the application of machine learning techniques to identify risk and protective behaviors among MSM using text data. By leveraging large language models, such as ChatGPT, and employing embedding vectors, contextual
information can be captured, resulting in more accurate representations of related texts. The study aims
to assess the accuracy with which different risk behaviors can be identified from text data and evaluate
the effectiveness of ChatGPT embedding as a novel
feature in this process.

To the best of our knowledge, the use of ChatGPT Embedding as a feature to predict harmful drinking or high-risk sexual behavior has not been previously explored in any earlier studies.\par
The primary contribution of this study lies in:\par
1- assessing the accuracy with which different risk behaviors can be identified from text data.\par
2- Assessing the accuracy and effectiveness of ChatGPT Embedding as a novel feature for identifying risk behaviors from text data.\par
3- Shedding light on the potential of machine learning-driven interventions to address the health disparities faced by MSM.\par
Through this pioneering research, we hope to pave the way for more efficient, personalized, and accessible interventions that can significantly impact the well-being of MSM and other vulnerable populations.

\section{Method}
\subsection{Study Design}

Data are derived from the uTECH Study, a three-arm randomized comparison trial designed to primarily assess the feasibility, acceptability, and appropriateness of the "uTECH" intervention. The overall study design, intervention framework, and cross-platform data collection procedures are described in detail elsewhere \cite{uTECHFramework}.
“uTECH” utilizes a machine-learning algorithm to deliver text message-based HIV prevention and substance use harm reduction messages for young sexual and gender minority individuals who have sex with men.
As secondary outcomes, the study aims to accurately identify risk and protective behaviors, including alcohol use and binge drinking, number of sexual partners, and PrEP uptake and retention.
Social media and dating app data were collected at participants’ baseline and 6- and 12-month follow-up sessions.
Self-reported behavioral data were collected as part of baseline, 3-month, 6-month, 9-month, and 12-month surveys.
During these surveys, participants completed the AUDIT-C to measure alcohol use and binge drinking and were also asked questions on number of recent sexual partners and current PrEP use (refer to Tables \ref{tab:survey1} and \ref{tab:survey2}).
The AUDIT-C questionnaire was graded from 0--12 points, categorized as follows: 0--3 points as Low Risk, 4--5 points as Moderate Risk, 6--7 points as High Risk, and 8--12 points as Severe Risk.
The questions included in the assessment covered the frequency of alcohol use, the typical number of drinks consumed, and the frequency of binge drinking (defined as having 6 or more drinks on one occasion).

\subsection{Eligibility}
To be eligible for the study, participants had to meet the following criteria: (1) be age 18 to 29, (2) identify as a sexual or gender minority, (3) have had anal or oral sex in the past 3 months, (4) have used substances (such as alcohol, marijuana, poppers, methamphetamines, heroin, cocaine, ecstasy, etc.) in the past 3 months, (5) have had sex while using substances in the past 3 months, (6) have a negative or unknown HIV status, (7) have used a dating app to meet sexual or substance use partners in the past 3 months, (8) own a smartphone, (9) reside in the United States, (10) be willing to participate in a 12- month study, and (11) be able to provide informed consent. The study procedures were reviewed and approved by UCLA South General IRB (IRB\#19-000805). 

\subsection{Participant Recruitment}
224 Participants were recruited through various methods, including online outreach (paid advertising on social media like Facebook, Instagram, Grindr, and Craigslist, and reaching out to community organizations and influential users) and in person outreach (distribution of recruitment materials at events and locations attended by LGBTQ+ such as sexual health fairs, community centers, substance use treatment centers, non-profit organizations, retail businesses, and Pride festivals).

\subsection{Screening}
Participants underwent an initial online screener survey hosted on the Qualtrics platform., which was used to determine eligibility and assess their willingness to provide social media data. If individuals met the initial criteria, additional screens were shown to collect contact information, including their smartphone’s phone number. Qualtrics data was cross-checked using specific websites to identify Voice-over Internet Protocol (VOIP) numbers, blacklisted IP addresses, or the use of Virtual Private Network (VPN) software, indicating potential fraud or non-US users by- passing geolocation restrictions. Participants meeting eligibility criteria, passing fraud checks, and completing the screeners were considered eligible. Study staff then contacted them via phone, email, or text message to schedule an onboarding session using Calendly, an online scheduling service. Automated reminders were sent once appointments were booked. All onboarding appointments took place over the Zoom conferencing platform

\subsection{Consent Process}
Consent was obtained from all participants during the onboarding and enrollment process conducted on Zoom. The interviewer utilized the "Share Screen" function to display the consent document, while verbally explaining its contents to the participant. The consent document outlined the collection of text data while explicitly stating that photos and videos would not be collected. Participants were also informed about the protection and storage of their data. If they agreed to participate, the interviewer recorded their consent, and a copy of the consent was emailed to them for their records.

\subsection{Data Collection }
Participants were asked to provide social networking app paradata from their dating apps (Grindr or Tinder) and social media accounts (Twitter, Instagram, Reddit, Snapchat, and Facebook) to send to the research team. The data requests occurred at baseline and 6-month follow-up.
The collected data was stored securely on BOX.
A Python script was developed to download the data from BOX to our server using the BOX API, and it was regularly updated on a biweekly basis. During this process, all the text files were downloaded recursively, excluding photos, videos, and voice messages, while downloading the remaining data.

The labels for the extracted data were derived from the wellness survey, comprising the Sexual Behavior and Substance Use Survey. The survey was adapted from the U.S. Centers for Disease Control and Prevention (CDC)’s HIV/Pre-exposure prophylaxis (PrEP) clinical practice guideline.\cite{Cen2} 
Furthermore, there was an additional survey focused on alcohol usage\cite{alcohol}.
We extracted Positive/Negative labels to indicate the presence or absence of specific behaviors based on the survey responses. 
The survey questionnaires can be found in Appendix.

\subsection{Data Extraction}

Data was extracted from downloaded files, including JSON, HTML, CSV, etc. Messages sent by participants, both public and private, were meticulously extracted, excluding received messages. The extracted data was stored in separate CSV files for each social or dating app, with three columns: user ID, message text, and date and time of transmission. To avoid duplicated data, entries with identical user IDs, dates, times, and message content were removed.

\subsection{Data Preprocessing}
From an initial pool of 224 iOS users, we excluded Facebook data due to its limited relevance and duplicated Grindr data to enhance prediction of risk behaviors. We retained texts no older than six months, aligning with our survey's timeframe. To ensure model robustness, we excluded user IDs with less than 30 days of data or fewer than 1000 messages. After these steps, our analysis focused on 160 users.

\subsection{Feature Extraction}
To apply machine learning techniques on the collected data, we manually extracted various features. These feature extraction techniques are described in the following subsections
\subsubsection{Riskwords}

The first set of featured comprising a total of 330 features, focused on words and phrases associated with drug use or sexual behaviors. To identify these words and phrases, a predefined phrase list was employed, which had previously been proven effective in identifying MSM HIV risk behavior and substance use in a prior study\cite{Ova}. The phrase list consisted of various categories, each containing specific words related to a particular type of substance use or sexual behavior. For each word in the list, the frequency of word usage was calculated as follows:

\[
\text{{freq}}(w) = \frac{{\text{{\# of days with word }} w}}{{\text{{\# of days with text data}}}}
\]

This calculation provided a measure of how frequently a particular word or phrase was used within the available text data.

\subsubsection{Riskword categories}
The second set of features consisted of 33 features, considered higher-level phrase categories and the frequency of using words corresponding to each category. 
We used our previously defined frequency formula to determine frequencies.
\subsubsection{BERT}
The third set of features was generated using the Bidirectional Encoder Representations from Transformers (BERT) language model \cite{devlin}. Specifically, we utilized a pretrained BERT model that was trained for sentiment analysis on Twitter data \cite{tweet}. We selected this model because Twitter data typically exhibits similar language patterns to other social media and messaging platforms. To generate the features, we removed the last fully connected layer of the model, allowing us to obtain text embeddings from the model. These text embeddings were applied to each individual day of data for each participant. Subsequently, the text embeddings were averaged across all days of data for each participant, resulting in a 768-dimensional feature vector for each participant.

\subsubsection{LIWC}
The fourth set of features was computed by Linguistic Inquiry and Word Count (LIWC) software \cite{LIWC}, which uses built-in dictionaries to capture social and psychological states. It computes features describing how much an individual talks about a variety of topics, such as money, physical intimacy, or leisure activities, and it also computes higher-level descriptive features to measure factors, such as analytical thinking, authenticity, and emotional tone. It has been used in numerous studies to, for example, analyze fake news \cite{fake}, social media posts \cite{Soc1,Soc2}, online reviews \cite{review}, and college admission essays \cite{Alv}. 
We used it to generate features for each individual message, and we calculated the average across all messages for each participant yielded in 118 features.

\subsubsection{ChatGPT Embedding}

The fifth set of features was generated using ChatGPT Embedding. When a text message is provided to ChatGPT, it converts the text into a vector embedding that captures the underlying notion or meaning behind the text. We utilized the OpenAI API for this purpose.

\vspace{.2cm}
\textbf{Naive Approach}:
\begin{enumerate}[itemsep=0pt]
    \item Appended all messages associated with a user ID.
    \item Fed the combined messages to the model ”text-embedding-ada-002” to retrieve the Embedding vector (length of 1536).
    \item Tokenized the text messages using the ”text-embedding-ada-002” tokenizer (”cl100k base”).
    \item Managed the token limit and rate of requests to fit the model's maximum input.
\end{enumerate}

\textbf{Shortcomings}:
\begin{itemize}[itemsep=0pt]
    \item Poor model performance on the aggregated features.
    \item Combining all texts initially likely compromised capturing the meaning of individual messages with pertinent details.
\end{itemize}

\textbf{Motivation for Improved Approach}:

The inability of the naive approach to effectively capture individual message meanings necessitated a new approach.

\textbf{Modified Approach}:

To address the limitations of the naive approach, we have implemented a modified algorithm for generating ChatGPT embeddings. This modified approach aims to improve the capture of meaningful information from individual messages and optimize the processing efficiency. The core of this approach is outlined in pseudo-code below:

\begin{algorithm}[H]
\small
\hrule
\caption{GPT}
\hrule
\begin{algorithmic}[1]

\Function \texttt{get\_embedding(result)}:
    \State  \texttt{outputs} = []

    \For {\texttt{text in result}}
        \State  \texttt{text\_embedding} = \texttt{OpenAI API(text)}
        \State  Append \texttt{text\_embedding} to \texttt{outputs}
    \EndFor
    \State \Return \texttt{outputs}
\EndFunction
\Function \texttt{num\_tokens(string)}:
    \State Return token count of \texttt{string} 
\EndFunction

\Function \texttt{join\_strings\_list(string\_list)}:
    \State Initialize empty lists: \texttt{joined\_l}
    \State Initialize empty lists: \texttt{current\_l}
    \State \texttt{current\_token\_count} = 0
    \For{\texttt{string in string\_list}}
        \State \texttt{string\_token\_c} = \texttt{num\_tokens(string)}
        
        \If {\texttt{current\_token\_c < token\_limit}}
            \State Append \texttt{string} to \texttt{current\_l}
            \State Update \texttt{current\_token\_c}
        \Else \State Append \texttt{current\_l} to \texttt{joined\_l}
        \State Reset \texttt{current\_list} and \texttt{current\_token\_c}
        \EndIf
    \EndFor
    \State  Append last \texttt{current\_l} to \texttt{joined\_l}
   
    \State \Return \texttt{joined\_string\_lists}
\EndFunction

\For {\texttt{each user\_id}}:
    \State \texttt{message\_list} = text data associated with \texttt{user\_id}
    \State  \texttt{Joined} = \texttt{join\_strings\_list(message\_list)}
    \State \texttt{GPT} =\texttt{mean(get\_embedding(Joined))}
\EndFor
\end{algorithmic}
\hrule
\end{algorithm}

\textbf{Rationale}:
\begin{itemize}[itemsep=0pt]
    \item Sending individual messages independently to the ChatGPT API captures the meaning of each message more effectively.
    \item Challenges faced due to the API's rate limits demanded a more efficient approach.
    \item Grouping messages into lists and sending as one request improved efficiency.
    \item Averaging vectors enabled capturing the overarching latent semantics behind all messages for a particular user.
\end{itemize}

\subsubsection{ChatGPT on risk Messages and Words}

A final feature set combines risk words and ChatGPT Embedding to address an issue with the previous feature set. With the previous approach, averaging all messages diminished the impact of messages containing risk words. To rectify this, we decided to focus solely on risk words
\vspace{.4cm}

\textbf{Naive Approach}:
\begin{enumerate}[itemsep=0pt]
    \item Identified all messages containing risk words.
    \item Used the preceding algorithm to extract the ChatGPT embedding for each of these messages.
    \item Averaged these numeric vectors. This average was then normalized by multiplying with the ratio of messages with risk words to the total messages for each user, emphasizing users with a higher proportion of risk messages.
\end{enumerate}

\textbf{Shortcomings}:
\begin{itemize}[itemsep=0pt]
    \item Extracting embeddings for complete messages was cumbersome and inefficient both in time and cost.
    \item The significance of risk words might be overshadowed when situated within extensive messages.
\end{itemize}

\textbf{Motivation for Improved Approach}:
The naive approach's focus on whole messages diluted the impact of risk words, prompting a more targeted refinement.

\textbf{Modified Approach}
To surmount the naive approach's shortcomings, a revised strategy was crafted to extract embeddings only from risk words, better representing user communication. The pseudo-code is outlined below:

\begin{algorithm}[H]
\small
\hrule
\caption{GPT\_riskW}
\hrule
\begin{algorithmic}[1]
\Function \texttt{riskW\_embedding(user\_messages)}:
    \For{\texttt{message} IN \texttt{user\_messages}}
        \For{\texttt{word} IN \texttt{message}}
            \If{\texttt{word} IS \texttt{risk}}
                \State Append \texttt{word} TO \texttt{riskW\_l}
            \EndIf
        \EndFor
    \EndFor
    \State \texttt{appended\_riskW} = Join \texttt{riskW\_l} WITH spaces
    \State \texttt{embedding} \texttt{ChatGPT\_Model(appended\_riskW)}
    \State \Return \texttt{embedding}
\EndFunction

\For{\texttt{each user}}
    \State \texttt{message\_list} = text data associated with \texttt{user\_id}
    \State \texttt{GPT\_riskW} = \texttt{riskW\_embedding(message\_list)}
\EndFor
\end{algorithmic}
\hrule
\end{algorithm}

\textbf{Rationale}:
\begin{itemize}[itemsep=0pt]
    \item \textbf{Significance \& Noise Reduction:} By directly focusing on the risk words, the embeddings become more representative of the sentiments these words carry, while simultaneously eliminating the noise introduced by other non-risk content in the messages.
    \item \textbf{Efficiency:} The approach is computationally less demanding by processing only the risk words, leading to gains in both time and cost.
    \item \textbf{Enhanced Similarity:} This method ensures users who use more risk words have embeddings that are more alike. Moreover, users who employ the same risk words will have even more congruent embeddings.
    \item \textbf{Improved Results:} Through experimentation and as evidenced in Tables \ref{m}, \ref{tab:Fisher} and \ref{tab:my_table} , this modified approach not only enhanced classification results but also provided more insightful features among all GPT-generated ones.
\end{itemize}

\subsection{Model Training}
Based on the survey data mentioned in the Data Collection section\cite{Cen2} \cite{alcohol}, participants who responded with "Decline to answer" or "I don't know" to a particular question had their specific answers excluded from both model training and evaluation. However, their other survey responses were still utilized for analysis and model development. To predict answers, individual classification models were trained for each question. Due to some questions having overlapping answer options, they were subdivided into distinct questions, and separate models were constructed for each subset. Certain questions had a limited number of positive responses, such as the case where no participants reported being in a substance use treatment. Consequently, the discussion primarily centered around four labels that were considered to be the most informative.
We employed Scikit-learn \cite{Scikit} for our classification tasks. Specifically, we trained a Logistic Regression model with a maximum of 1000 iterations for its simplicity and interpretability \cite{reg}, a Gradient Boosting Classifier with 100 boosting stages to capture more intricate data relationships \cite{gboost}, and a Support Vector Classifier (SVC) with a linear kernel using default hyperparameters for a balance between performance and interpretability \cite{SVM}.These models were chosen to represent linear models as well as a more complex non-linear model.

\subsubsection{Naive Approach}
\begin{itemize}[itemsep=0pt]
    \item Trained separate models using each feature category, as detailed in the feature extraction section.
    \item Evaluated combinations of these features, prioritizing combinations believed to enhance prediction.
    \item Used leave-one-out cross-validation due to the limited number of participants.
\end{itemize}

\textbf{Shortcomings}:
\begin{itemize}[itemsep=0pt]
    \item The predictive performance did not meet our expectations.
    \item The model possibly overfitted the data.
    \item There was no systematic method to select the most informative features.
\end{itemize}

\textbf{Motivation for Improved Approach}:
The suboptimal results from the naive approach prompted a reevaluation of our strategy, particularly in terms of feature selection, to mitigate potential overfitting and enhance model performance.

\subsubsection{Improved Approach}

Fisher's score is a foundational metric for feature selection in classification challenges. It quantifies the relevance of features by evaluating their ability to differentiate between classes. Higher scores signify more significant features. For a dataset with $n$ points, $c$ classes, and $d$ features, where $X$ is the feature matrix (shape $n \times d$) and $y$ is the target vector (shape $n$), the score for feature $i$ ($0 \leq i < d$) is:
\[
\text{Fisher's Score}(i) = \frac{\text{BCV}(i)}{\text{WCV}(i)}
\]
where:
\begin{align*}
\text{BCV}(i) &= \sum_{j=0}^{c-1} n_j \cdot (\text{Class Mean}_j(i) - \text{Overall Mean}(i))^2 \\
\text{WCV}(i) &= \sum_{j=0}^{c-1} \sum_{x \in \text{Class}(j)} (x_i - \text{Class Mean}_j(i))^2
\end{align*}
$\text{Class Mean}_j(i)$ is the mean of feature $i$ for class $j$, $\text{Overall Mean}(i)$ is the feature's mean across all points, and $n_j$ is the number in class $j$.

The approach is further elucidated in the pseudo-
code below
\begin{algorithm}[H]
\small
\caption{Feature Selection using Fisher Scores with Divide and Conquer}
\begin{algorithmic}[1]

\Function \texttt{ComputeScore(f, lbl)}:
  \State Extract \texttt{Between-Class Variance (BCV)}
  \State Extract \texttt{Within-Class Variance (WCV)}
  \State \texttt{Score = BCV / WCV} 
  \State \Return \texttt{Score}
\EndFunction

\Function \texttt{DACSearch(Data, label)}:
  \State Initialize \texttt{bestK = 1} and \texttt{bestF1 = 0}
  
  \State \textbf{Coarse Search}
  \For {each \texttt{k} in range(1, 160, 20)}:
    \State Get \texttt{topK} features using \texttt{ComputeScore}
    \State Compute average \texttt{F1} over 5-folds with \texttt{topK} features
    \State Update \texttt{bestK} if new \texttt{F1} is better
  \EndFor
  
  \State \textbf{Fine Search} around \texttt{bestK}
  \For {each \texttt{k} in range(bestK-10, bestK+10, 5)}:
    \State Repeat the above steps within this range
  \EndFor
  
  \State \Return \texttt{bestK}
\EndFunction

\State Initialize arrays \texttt{all\_predictions = []} and \texttt{all\_true\_labels = []}

\For {each instance in the dataset}:
  \State Hold out the current instance as \texttt{t\_inst} and its label as \texttt{t\_lbl}
  \State \texttt{bestK = DACSearch(data - t\_inst, labels - t\_lbl)}
  \State Train final model on entire data excluding \texttt{t\_inst} with top \texttt{bestK} features
  \State Predict for \texttt{t\_inst} and add prediction to \texttt{all\_predictions}
  \State Add \texttt{t\_lbl} to \texttt{all\_true\_labels}
\EndFor

\State Compute and report F1 score using \texttt{all\_predictions} and \texttt{all\_true\_labels}

\end{algorithmic}
\end{algorithm}

\textbf{Rationale}:
\begin{itemize}
   
    \item Implementing the Fisher score selection enabled us to focus on the most informative features, hence, efficiently preventing overfitting. The rationale behind using the Fisher score is its intrinsic ability to measure the discriminative power of individual features based on their variances.
    \item To avoid any form of data leakage and provide an unbiased assessment of the performance, for every iteration, one instance was held out from the dataset prior to calculating the Fisher score on the training data.
    \item Furthermore, the iterative determination of the optimal \( K \) value for feature selection was executed with a divide-and-conquer approach. This ensured that we not only covered the search space efficiently but also homed in on the optimal \( K \) value more accurately.
    \item Emphasizing on the divide-and-conquer method, we commenced with a broad interval coarse search for \( K \) values, and subsequently narrowed down our search, refining into smaller intervals. This two-step search provided a good balance between computational efficiency and finding the best \( K \) value.
\end{itemize}

\textbf{Outcome}:
Utilizing this advanced methodology countered the overfitting issue observed in the naive approach. It allowed us to emphasize the most informative features, yielding a markedly improved predictive performance.

\section{Result}
\subsection{Data Statistics}

We collected data from 224 iOS users and decided to exclude Facebook data due to its limited relevance for predicting risk behaviors. Conversely, we duplicated data from Grindr, expecting it to provide valuable insights into risk behaviors. Our experimentation confirmed that this approach improved the prediction task. For detailed breakdown of number of messages that used in model after filtering, see Table \ref{tab:app_usage}.

\begin{table*}[h]
\begin{threeparttable}
\small
\centering
\begin{tabularx}{\textwidth}{l*{7}{X}}
\toprule
& \multicolumn{3}{c}{\textbf{Dating apps}} & \multicolumn{4}{c}{\textbf{Social apps}} \\
\cmidrule(lr){2-4} \cmidrule(lr){5-8}
Total & Grindr  & Grindr\_Pn\tnote{*} & Tinder & Instagram & Snapchat & Twitter & Reddit \\
\midrule
1,316,699 & 842,284 & 374 & 11,109 & 409,235 & 50,622 & 2,623 & 452 \\
\bottomrule
\end{tabularx}
\begin{tablenotes}
      \item[*] \scriptsize{Grindr Profile notes}
    \end{tablenotes}
   \end{threeparttable}
\caption{\small{Summary of app messages considered in the model }} \label{tab:app_usage}
\end{table*}

\begin{table*}[!t]
\small
\centering
\begin{tabular}{cccccc}
\hline
Total Messages & Mean & Median & Maximum & Minimum & SD\\
\hline
1,316,699 & 8,229.37 & 4,937.0 & 47,050 & 1,016 & 8,508.54 \\
\hline
\end{tabular}
\caption{\small{Statistics of messages considered in the model}}
\label{tab:Statistics}
\end{table*}

Upon consolidating the data and applying filtering criteria, we were left with 160 users who met the filtering requirements. Table \ref{tab:Statistics} contains the statistics for the number of messages used in our models, including the maximum and minimum values, which represent the highest and lowest number of messages sent by a single user ID.

\begin{table*}[!t]
\small
\centering

 \begin{threeparttable}
\label{tab:data}
\begin{tabular}{lccc}
\hline
\multirow{2}{*}{} & \multirow{2}{*}{Positive} & \multirow{2}{*}{Negative} & \multirow{2}{*}{Total} \\

Alcohol-related problem & & & \\
Binge Monthly & 82 (49\%) & 73 (51\%) & 155 \\
AUDIT-C High\tnote{*} & 53 (35\%) & 97 (65\%) & 150 \\
\\

Sexual behavior & & & \\
Over 5 partners & 84 (52.5\%) & 76 (47.5\%) & 160 \\
\\

Protective behaviors & & & \\
Takes PrEP & 92 (57.5\%) & 68 (42.5\%) & 160 \\ \hline
\end{tabular}
  \begin{tablenotes}
      \item[*] \scriptsize{The AUDIT-C score, derived from the Alcohol Use Survey, is considered high when it equals or exceeds 6.}
    \end{tablenotes}
  \end{threeparttable}
\caption{\small{Labels count}}
\label{tab:labels}
\end{table*}

\begin{table*}[!t]
\small
\centering

\begin{tabular}{lcc}
\hline
\multirow{2}{*}{} & \multirow{2}{*}{Enhanced Approach} & \multirow{2}{*}{Naive Approach}  \\

Alcohol-related problem & & \\
\cline{2-3}
Binge Monthly & .78 &  .64 \\
AUDIT-C High\tnote{*} & .63 & .44 \\
\cline{2-3}
Sexual behavior & &  \\
\cline{2-3}
Over 5 partners & .78  &  .66 \\

\cline{2-3}
Protective behaviors & &  \\
\cline{2-3}
Takes PrEP & .64 &  .62 \\ \hline
\end{tabular}
  \caption{\small{compare the best results for each label based on F1 for minority class.}}
\label{tab:compare}
\end{table*}

Based on the survey data mentioned in the Data Collection section\cite{Cen2} \cite{alcohol}, participants who responded with "Decline to answer" or "I don't know" to a particular question had their specific answers excluded from both model training and evaluation. However, their other survey responses were still utilized for analysis and model development. To predict answers, individual classification models were trained for each question. Due to some questions having overlapping answer options, they were subdivided into distinct questions, and separate models were constructed for each subset. Certain questions had a limited number of positive responses, such as the case where no participants reported being in a substance use treatment. Consequently, the discussion primarily centered around four labels that were considered to be the most informative. In the breakdown presented in Table \ref{tab:labels}, you can find the distribution of positive and negative labels for these four categories, which served as the basis for training the models.

\subsection{Model profermances}
The outcomes of survey response prediction using individual feature types and their combinations are presented in Table \ref{m}. Feature combinations were selected both based on their individual results and based on whether they were presumed to provide non-overlapping information. For example, we avoided combining highly correlated features groups, such as risk words and risk word Categories. Also, we decided to exclude GPT on risk messages from feature combination because it demonstrated poor individual performance in compare to GPT solely on risk word. Our initial model combining GPT on risk messages with other features confirmed our hypothesis that using it in combinations would not make further advantage in the prediction task.

In the improved method that incorporates feature selection, we kept only the K first features with largest Fisher scores. The best results we could get for minority class for each labels for Binge monthly, having more than 5 partners, AUDIT-C High and Taking PreP was F1 socre 0.78 , 0.78 , 0.64 , 0.63 respectively. All of these top results were obtained using SVM with linear kerner. You can find a camparison of the best result from naive approach and improved approach in Table \ref{tab:compare}.

\subsection{Identify best sets of features}
In the leave-one-out method, Since each time, the set of features might be different we found the average number of features used in the model from different sets. The features generated using ChatGPT on risk words contributed the most to the improved method (Table \ref{tab:Fisher})

To further assessed the information value of different feature subsets, we computed correlations between each feature and the target labels. Features with correlations above specific thresholds (0.2) were retained.\newline
We have also performed t-tests to identify the most relevant set of features. Grouping the data based on the binary target variable, we calculated p-values using scipy.stats' ttest\_ind function. Features with p-values less than 0.05 were considered relevant.
The study's findings, as presented in Table VI, showcase the significant impact of different feature sets used in both methods. Notably, employing ChatGPT solely on risk words emerges as the most influential in generating relevant features across various labels. This approach notably outperforms others in nearly all categories, with the sole exception being in the context of 'Takes PrEP'. This evidence underscores the substantial utility of ChatGPT, particularly highlighting its effectiveness in embedding for classification tasks. ChatGPT's capability to provide insightful contributions affirms its value in enhancing the accuracy and depth of analysis in such studies

\begin{table*}[!h]
    \small
    \centering
     \begin{threeparttable}
    \begin{tabular}{ccccccccc}
        \hline
         & RiskWord & Riskcat & BERT & LIWC & GPT\tnote{1}& GPT\_RiskM\tnote{2} & GPT\_RiskW\tnote{3} & K \\
        \hline
        Binge Monthly & 0 & 0 & 0 & 2.92 & 8.93 & 4.29 & 18.86 & 35 \\
        \hline
        AUDIT-C High  & 0 & 0 & 0 & 0.02 & 11.9 & 2.99 & 10.09 & 25\\
        \hline
        Over 5 partners & 0 & 0 & 0 & 1.01 & 3.06 & 1.99 & 8.94 & 15 \\
        \hline
        Takes PrEP & 0 & 0 & 37 & 0.13 & 0.02 & 1.02 & 31.83 & 70\\
        \hline
    \end{tabular}
    \begin{tablenotes}
      \item[1] \scriptsize{Using ChatGPT Embeddings on all messages}
      \item[2] \scriptsize{Using ChatGPT Embeddings on messages with risk words}
      \item[3] \scriptsize{Using ChatGPT Embeddings on risk words}
    \end{tablenotes}
     \end{threeparttable}
    \caption{\small{Average contribution of different set picked by Fisher score in the final model}}
    \label{tab:Fisher}
\end{table*}

\begin{table*}[!h]
    \small
    \centering
   
    \begin{tabular}{ccccccccc}
        \hline
        \multirow{2}{*}{} & \multicolumn{8}{c}{\textbf{using correlation $>$ 0.2}} \\
        \cline{2-9}
         &  RiskWord & Riskcat & BERT & LIWC & GPT & GPT\_RiskM & GPT\_RiskW & Total \\
        \hline
        Binge Monthly & 1 & 0 & 2 & 9 & 35 & 16 & 65 & 128\\
        \hline
        AUDIT-C High & 0 & 0 & 1 & 1 & 39 & 7 & 45 & 93\\
        \hline
        Over 5 partners & 1 & 0 & 6 & 1 & 33 & 19 & 101 & 161\\
        \hline
        Takes PrEP & 2 & 1 & 139 & 5 & 16 & 26 & 152 & 341\\
        \hline
        \multirow{2}{*}{} & \multicolumn{8}{c}{\textbf{using t-test p-value $<$ 0.05}} \\
        \cline{2-9}
         & RiskWord & Riskcat & BERT & LIWC & GPT& GPT\_RiskM & GPT\_RiskW & Total \\
        \hline
        Binge Monthly & 8 & 0 & 8 & 16 & 137 & 83 & 172 & 424\\
        \hline
        AUDIT-C High  & 4 & 0 & 2 & 3 & 92 & 36 & 119 & 256 \\
        \hline
        Over 5 partners & 10 & 3 & 60 & 11 & 134 & 97 & 247 & 562 \\
        \hline
        Takes PrEP & 12 & 3 & 234 & 11 & 67 & 92 & 318 & 737 \\
        \hline
    \end{tabular}
    
    \caption{\small{Contribution of different sets of features in the most correlated and relevant features}}
    \label{tab:my_table}
\end{table*}

\clearpage

\begin{sidewaystable}

\small
  \vspace{8.5cm}
   \begin{threeparttable}
  \begin{subtable}{\textwidth} 
  \begin{tabular}{|cccccccc|}
   
    \hline
    
    Labels & RiskWord & Riskcat & BERT & LIWC & GPT & GPT\_RiskM & GPT\_RiskW  \\
    \hline
    Alcohol-related problem & & & & & & &  \\ 
    \hline
    Binge Monthly & .49 - .43 - .52 & .4 - .53 - .32 &  .54 - .54 - .53 & .58 - .55 - .56 & .6 - .56 - .56 & .49 - .43 - .49 & .53 - .58 - .54 \\
    \hline
    AUDIT-C High & .31 - .19 - .42 & .29 - .2 - .29  & .4 - .22 - .37 & .39 - .33 - .39 & .35 - .2 - .34 & .24 - .22 - .27 & .39 - .31 - .4 \\
    \hline
    Sexual behavior & & & & & & & \\ 
    \hline
    Over 5 partners & .56 - .54 - .57 & .49 - .55 - .39 & .43 - .49 - .42  & .5 - .37 - .5 & .49 - .49 - .44 & .4 - .42 - .42 & .54 - .54 - .51 \\
    \hline
    Protective behaviors & . & & & & & & \\ 
    \hline
    Takes PrEP & .45 - .49 - .32 & .46 - .4 - .47 & .53 - .51 - .47 & .42 - .39 - .5 & .34 - .37 - .39 & .33 - .4 - .31 & .59 - .59 - .56\\
    \hline
  \end{tabular}
  \end{subtable}

  \vspace{1cm}
  
   \begin{subtable}{\textwidth} 

  \begin{tabular}{|cccccccc|}
    \hline
    
    Labels &   \parbox{2cm}{\centering LIWC, BERT} & \parbox{2cm}{\centering BERT, GPT} & \parbox{2cm}{\centering LIWC, GPT} & \parbox{2cm}{\centering GPT GPT\_RiskW} &  \parbox{2cm}{\centering Riskword, GPT} & \parbox{2cm}{\centering Riskword BERT} & \parbox{2cm}{\centering RiskCat GPT\_RiskW}   \\
    \hline
    Alcohol-related problem & & & & & & & \\ 
    \hline
    Binge Monthly & .55 - .48 - .58 & .57 - .64 - .57 & .63 - .59 - .6 &  .53 - .58 - .56 & .5 - .56 - .5 & .53 - .53 - .54 & .49 - .54 - .51 \\
    \hline
   AUDIT-C High &  .35 - .24 - .38 & .31 - .19 - .36 & .32 - .24 - .37 & .41 - .33 - .4 & .27 - .27 - .31 & .44 - .2 - .43 & .39 - .31 - .38  \\
    \hline
    Sexual behavior & & & & & &  &  \\ 
    \hline
    Over 5 partners & .53 - .47 - .5 & .54 - .57 - .51 & .49 - .51 - .47 & .6 - .52 - .58 & .47 - .51 - .43 & .52 - .49 - .51 & .54 - .56 - .55 \\
    \hline
    Protective behaviors & & & & & & &  \\ 
    \hline
    Takes PrEP & .5 - .45 - .5 &  .5 - .57 - .49 & .43 - .41 - .45 &  .53 - .56 - .54 & .37 - .43 - .36 & .47 - .52 - .47 & .57 - .61 - .58 \\
    \hline
   
  \end{tabular}
  \end{subtable}
  
\vspace{1cm}

   \begin{subtable}{\textwidth} 
  \begin{tabular}{|cccccccc|}
    \hline
    
    Labels  & \parbox{2cm}{\centering Riskword GPT\_RiskW} & \parbox{2cm}{\centering GPT, LIWC BERT} &  \parbox{2cm}{\centering BERT, GPT Riskword} & \parbox{2cm}{\centering GPT, BERT GPT\_RiskW} & \parbox{2cm}{\centering GPT, LIWC BERT GPT\_RiskW} &  \parbox{2cm}{\centering LIWC, GPT BERT Riskword} &  \parbox{2cm}{\centering LIWC, GPT BERT RiskCat}  \\
    \hline
    Alcohol-related problem & & & & & & & \\ 
    \hline
    Binge Monthly & .49 - .53 - .49 & .6 - .63 -.58 & .52 - .63 - .53 & .54 - .62 - .55 & .54 - .63 - .52 & .55 - .61 - .54 & .59 - .63 - .58 \\
    \hline
    AUDIT-C High &  .38 - .32 - .39 & .28 - .15 - .32 & .29 - .24 - .34 & .4 - .33 - .4 & .4 - .31 - .39 & .27 - .21 - .3 & .28 - .18 - .32 \\
    \hline
    Sexual behavior & & & & & & & \\ 
    \hline
    Over 5 partners & .56 - .61 - .55 & .57 - .52 - .58 & .54 - .59 - .5 & .6 - .66 - .61  & .61 - .63 - .6 & .58 - .58 - .53 & .55 - .55 - .56 \\
    \hline
    Protective behaviors & & & & & & & \\ 
    \hline
    Takes PrEP &  .55 - .6 -.56 &  .52 - .43 - .53 & .53 - .55 - .5 & .55 - .57 - .55 & .52 - .62 - .52 & .56 - .52 - .53  & .52 - .61 - .55 \\
    \hline
  \end{tabular}
  \end{subtable}

  \caption{\small{F1 scores for predicting answers to survey questions. F1 score was calculated for the less frequent response, which in most cases was the positive answer (answer frequencies are shown in Table \ref{tab:labels}). First value shows the score using logistic regression, the second value shows the score using a gradient boosting classifier and the third value shows the score using SVM with linear kernel}}
  \label{m}
  \end{threeparttable}
 
\end{sidewaystable}
\clearpage

\section{Discussion}

\subsection{Comparison with Prior Work}

Prior studies have explored the use of passively collected smartphone data and machine learning to infer HIV-related risk and protective behaviors among MSM using Android-based sensing platforms \cite{AndroidRisk,AndroidPrEP}. These efforts demonstrated the feasibility of predicting outcomes such as substance use, sexual risk, and PrEP use from mobile text and behavioral features. In contrast, the present work focuses on iOS users and leverages a different data access and collection paradigm based on user-provided social media and dating app data.

The other similar work to ours is \cite{31} which identified HIV as well as methamphetamine and tetrahydrocannabinol (THC) use from social media messaging data. Our work differs from this targets that we predicted and also using newer LLM methods like chatGPT Embeddings.

Another similar study is \cite{45} which predicted alcohol, tobacco, prescription drug, and illegal drug use from Instagram data. They were able to detect alcohol use with statistical significance with (F1 score 72.4\%). In our study we could predict Binge monthly with a F1-score of 0.78 for negative class and 0.8 for positive class using SVM with linear kernel, thanks to our feature extraction and selection methods.

\subsection{Principal Results}

\begin{itemize}[itemsep=0pt]
    \item Our findings indicate that the text data collected from Social apps
and dating apps can predict behaviors such as monthly binge drinking, AUDIT-C high, having multiple sexual partners and the use of PrEP with high accuracy. 

    \item Our study has illustrated the utility of ChatGPT Embeddings in text mining classification. These embeddings capture valuable underlying information in vectors, resulting in improved predictive accuracy specifically when combining chatGPT features with traditional techniques based on predetermined word lists frequencies and language models like BERT and LIWC.
    \item To fully explore the power of ChatGPT Embeddings, additional analysis on larger datasets may unveil even more potential.

\end{itemize}

\subsection{Limitations}
Our study encountered several limitations. 
\newline Firstly, due to the restricted rules imposed by iPhone, we were limited to collecting data from social media and dating apps only. To gain deeper insights into participants' behaviors, it would have been beneficial to collect additional text data, such as their internet search history. The data collection was restricted to text data, with consent obtained solely for this purpose. However, Incorporating image data exchanged between individuals could have significantly enhanced our predictive capabilities. For instance, running NSFW models on images might have improved the accuracy of predicting sexual behaviors.

Secondly, some of the outcomes we aimed to predict were exceptionally infrequent, rendering the task nearly impossible. For instance, only a few participants reported engaging in daily binge drinking. The limited occurrence of these outcomes posed challenges for effectively training and evaluating machine learning models. Consequently, we shifted our focus to questions with a more reasonable distribution of both positive and negative responses
\section{Conclusion}
 In this study, we have demonstrated that Social apps and dating apps text data can be utilized to identify certain types of harmful drinking, high-risk sexual behavior, and protective actions effectively.
 These results highlight the potential of using text data to personalize interventions for individuals at high-risk, thereby reducing the burden of participating in intervention programs. However, further research is required to evaluate the effectiveness of interventions based on social media text data.
 Also we have shown that chatGPT Embedding can be very valuable and informative in prediction task.


\section*{Acknowledgment}

This study was funded by a grant from the National Insitute on Drug Abuse (1DP2DA049296-01). Cory J. Cascalheira completed work on this study while supported as a RISE Fellow with the National Institutes of Health (R25GM061222)
\bibliographystyle{IEEEtran}
\bibliography{myrefrences}

\clearpage
\section*{Appendix: Surveys}
\label{Appendix}
\begin{table}[htb]
\centering
\captionsetup[table]{skip=10pt}
\scriptsize
\setlength{\tabcolsep}{4pt} 
\begin{tabular}{ | p{4cm} | p{4cm}| } 
\hline
\textbf{Question} & \textbf{Answer options} \\
\hline
How old are you today? & 
1.	(Enter number)
 \\
\hline
In the past 3 months, have you used substances such as crystal meth or injectable drugs not prescribed to you by a physician? & 
1. Yes, methamphetamines\newline
2. Yes, injectable drugs not prescribed\newline
3.	Yes, used both\newline
4.	Neither\newline
5.	Decline to answer
 \\
\hline
In the past 3 months, were you in a substance use in-patient or out-patient treatment program? & 1.	Yes\newline
2.	No\newline
3.	Decline to answer\newline
 \\
 \hline
In the past 3 months, did you inject cocaine? & 1.	Yes\newline
2.	No\newline
3.	Decline to answer\newline
 \\
 \hline
In the past 3 months, did you inject methamphetamines? & 1.	Yes\newline
2.	No\newline
3.	Decline to answer\newline
 \\
 \hline
In the past 3 months, did you share injection equipment? & 1.	Yes\newline
2.	No\newline
3.	Decline to answer\newline
 \\
 \hline
In the past 3 months, did you inject in a group setting? & 1.	Yes\newline
2.	No\newline
3.	Decline to answer\newline
 \\
 \hline
Do you currently take PrEP? PrEP stands for pre-exposure prophylaxis and it is a medication that helps prevent HIV transmission. & 1.	Yes\newline
2.	No\newline
3.	Decline to answer\newline
 \\
  \hline
In the last 3 months, how many men have you had sex with? &1.$>$10\newline
2.	6-10\newline
3.	1-5\newline
4.	0\newline
5.	Decline to answer

 \\
  \hline
In the last 3 months, how many times did you have receptive anal sex (you were the bottom) with a man when he did not use a condom? &1. 1 or more times\newline
2.	0 times\newline
3.	Decline to answer

 \\
  \hline
In the last 3 months, how many of your male sex partners were HIV-positive? & 1.	More than 1 HIV+ male partners\newline
2.	1 HIV+ male partner\newline
3.	0\newline
4.	Don’t know\newline
5.	Decline to answer

 \\
  \hline
In the last 3 months, how many times did you have insertive anal sex (you were the top) with a man who was HIV-positive when you did not use a condom? & 1.	1.	5 or more times\newline
2.	0-4 times\newline
3.	Decline to answer

 \\
\hline
\end{tabular}
\caption{Sexual behavior and substance use survey}
\label{tab:survey1}
\end{table}

\begin{table}[t]
\centering
\scriptsize
\setlength{\tabcolsep}{4pt} 
\begin{tabular}{ | p{3cm} | p{3cm}| } 
\hline
\textbf{Question} & \textbf{Answer options} \\
\hline
In the past 6 months, how often did you have a standard drink containing alcohol? & 
1. Never\newline
2. Monthly or less \newline
3. 2 to 4 times a month\newline
4. 2 to 3 times a week\newline
5. four or more times a week\newline
 \\
\hline
How many standard drinks containing alcohol do you have on a typical day?  & 1. 1-2 \newline
2. 3-4\newline
3. 5-6\newline
4. 7-9\newline
5. 10 or more
 \\
 \hline
During the past 6 months, how often did you have 6 or more standard drinks on one occasion?(Binge drinking) &  1. Never\newline
2. Less than monthly\newline
3. monthly\newline
4. weekly\newline
5. daily or almost daily
 \\
 \hline

\end{tabular}
\caption{Alcohol use survey}
\label{tab:survey2}
\end{table}

\end{document}